\documentclass[]{article}
\usepackage{colacl}
\usepackage{graphicx}

\title{Advances in domain independent linear text segmentation}

\author{Freddy Y. Y. Choi\\
Artificial Intelligence Group\\
Department of Computer Science\\
University of Manchester\\
Manchester, England\\
{\it choif@cs.man.ac.uk}
}

\begin{document}
\maketitle

\begin{abstract}
This paper describes a method for linear text segmentation which is twice as accurate and over seven times as fast as the state-of-the-art \cite{reynar_1998}. Inter-sentence similarity is replaced by rank in the local context. Boundary locations are discovered by divisive clustering.
\end{abstract}

\section{Introduction}
Even moderately long documents typically address several topics or different aspects of the same topic. The aim of linear text segmentation is to discover the topic boundaries. The uses of this procedure include information retrieval \cite{hearst_plaunt_1993,hearst_1994,yaari_1997,reynar_1999}, summarization \cite{reynar_1998}, text understanding, anaphora resolution \cite{kozima_1993}, language modelling \cite{morris_hirst_1991,beeferman_et_al_1997} and improving document navigation for the visually disabled \cite{choi_2000iee}.

This paper focuses on domain independent methods for segmenting written text. We present a new algorithm that builds on previous work by Reynar \cite{reynar_1998,reynar_1994}. The primary distinction of our method is the use of a ranking scheme and the cosine similarity measure \cite{rijsbergen_1979} in formulating the similarity matrix. We propose that the similarity values of short text segments is statistically insignificant. Thus, one can only rely on their order, or rank, for clustering.

\section{Background}
Existing work falls into one of two categories, lexical cohesion methods and multi-source methods \cite{yaari_1997}. The former stem from the work of Halliday and Hasan \cite{halliday_hasan_1976}. They proposed that text segments with similar vocabulary are likely to be part of a coherent topic segment. Implementations of this idea use word stem repetition \cite{youmans_1991,reynar_1994,ponte_croft_1997}, context vectors \cite{hearst_1994,yaari_1997,kaufmann_1999,eichmann_et_al_1999}, entity repetition \cite{kan_et_al_1998}, semantic similarity \cite{morris_hirst_1991,kozima_1993}, word distance model \cite{beeferman_et_al_1997b} and word frequency model \cite{reynar_1999} to detect cohesion. Methods for finding the topic boundaries include sliding window \cite{hearst_1994}, lexical chains \cite{morris_1988,kan_et_al_1998}, dynamic programming \cite{ponte_croft_1997,heinonen_1998}, agglomerative clustering \cite{yaari_1997} and divisive clustering \cite{reynar_1994}. Lexical cohesion methods are typically used for segmenting written text in a collection to improve information retrieval \cite{hearst_1994,reynar_1998}. 

Multi-source methods combine lexical cohesion with other indicators of topic shift such as cue phrases, prosodic features, reference, syntax and lexical attraction \cite{beeferman_et_al_1997b} using decision trees \cite{miike_et_al_1994,kurohashi_nagao_1994,litman_passonneau_1995} and probabilistic models \cite{beeferman_et_al_1997,hajime_1997,reynar_1998}. Work in this area is largely motivated by the topic detection and tracking (TDT) initiative \cite{allan_et_al_1998}. The focus is on the segmentation of transcribed spoken text and broadcast news stories where the presentation format and regular cues can be exploited to improve accuracy.



\section{Algorithm}
Our segmentation algorithm takes a list of tokenized sentences as input. A tokenizer \cite{grefenstette_1994} and a sentence boundary disambiguation algorithm \cite{palmer_hearst_1994,reynar_ratnaparkhi_1997} or EAGLE \cite{reynar_1997b} may be used to convert a plain text document into the acceptable input format.

\subsection{Similarity measure}
Punctuation and uninformative words are removed from each sentence using a simple regular expression pattern matcher and a stopword list. A stemming algorithm \cite{porter_1980} is then applied to the remaining tokens to obtain the word stems. A dictionary of word stem frequencies is constructed for each sentence. This is represented as a vector of frequency counts.

Let $f_{i,j}$ denote the frequency of word $j$ in sentence $i$. The similarity between a pair of sentences $x,y$ is computed using the cosine measure as shown in equation \ref{eq:cosine}. This is applied to all sentence pairs to generate a similarity matrix.
\begin{equation}
sim(x,y) = \frac{\sum_{j} f_{x,j} \times f_{y,j}}{\sqrt{{\sum_{j} f_{x,j}^2} \times {\sum_{j} f_{y,j}^2}}}
\label{eq:cosine}
\end{equation}

\begin{figure}[h]
\begin{center}
\includegraphics[width=0.3\textwidth]{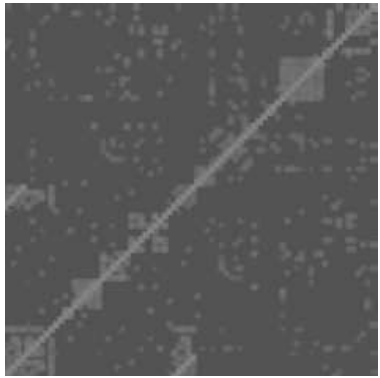}
\end{center}
\caption{An example similarity matrix.}
\label{fig:simmatrix}
\end{figure}

Figure \ref{fig:simmatrix} shows an example of a similarity matrix\footnote{The contrast of the image has been adjusted to highlight the image features. }. High similarity values are represented by bright pixels. The bottom-left and top-right pixel show the self-similarity for the first and last sentence, respectively. Notice the matrix is symmetric and contains bright square regions along the diagonal. These regions represent cohesive text segments.

\subsection{Ranking}
For short text segments, the absolute value of $sim(x,y)$ is unreliable. An additional occurrence of a common word (reflected in the numerator) causes a disproportionate increase in $sim(x,y)$ unless the denominator (related to segment length) is large. Thus, in the context of text segmentation where a segment has typically $< 100$ informative tokens, one can only use the metric to estimate the order of similarity between sentences, e.g. $a$ is more similar to $b$ than $c$.

Furthermore, language usage varies throughout a document. For instance, the introduction section of a document is less cohesive than a section which is about a particular topic. Consequently, it is inappropriate to directly compare the similarity values from different regions of the similarity matrix.

In non-parametric statistical analysis, one compares the rank of data sets when the qualitative behaviour is similar but the absolute quantities are unreliable. We present a ranking scheme which is an adaptation of that described in \cite{oneil_denos_1992}.

\begin{figure}[h]
\begin{center}
\includegraphics[width=0.48\textwidth]{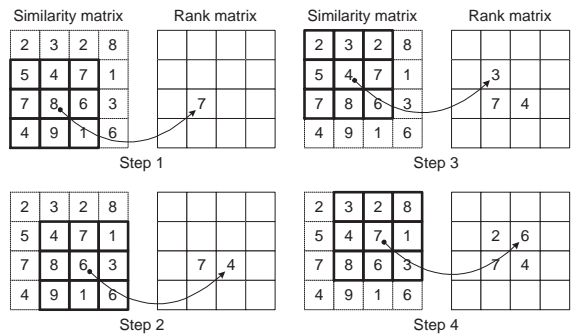}
\end{center}
\caption{A working example of image ranking.}
\label{fig:rank}
\end{figure}

Each value in the similarity matrix is replaced by its rank in the local region. The rank is the number of neighbouring elements with a lower similarity value. Figure \ref{fig:rank} shows an example of image ranking using a $3 \times 3$ rank mask with output range $\{0,8\}$. For segmentation, we used a $11 \times 11$ rank mask. The output is expressed as a ratio $r$ (equation \ref{eq:rankRatio}) to circumvent normalisation problems (consider the cases when the rank mask is not contained in the image).

\begin{equation}
r = \frac{\textrm{\# of elements with a lower value}}{\textrm{\# of elements examined}}
\label{eq:rankRatio}
\end{equation}

\begin{figure}[h]
\begin{center}
\includegraphics[width=0.3\textwidth]{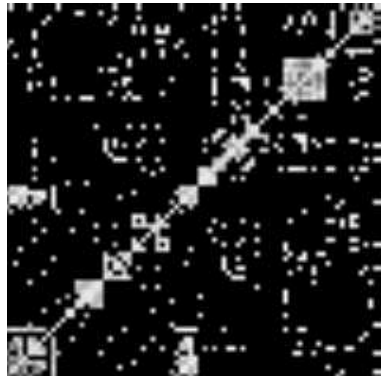}
\end{center}
\caption{The matrix in figure \ref{fig:simmatrix} after ranking.}
\label{fig:rankmatrix}
\end{figure}

\begin{figure}[h]
\begin{center}
\includegraphics[width=0.48\textwidth]{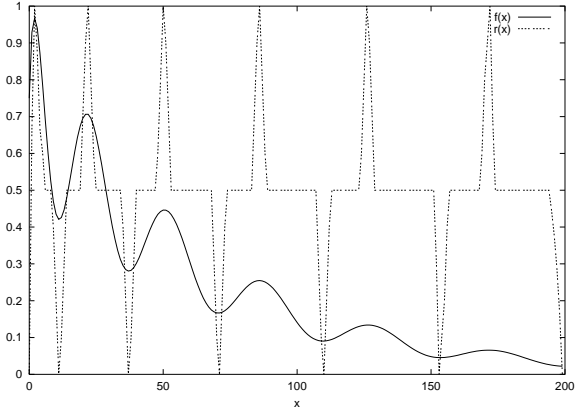}
\end{center}
\caption{An illustration of the more subtle effects of our ranking scheme.}
\label{fig:rankillus}
\end{figure}

To demonstrate the effect of image ranking, the process was applied to the matrix shown in figure \ref{fig:simmatrix} to produce figure \ref{fig:rankmatrix}\footnote{The process was applied to the original matrix, prior to contrast enhancement. The output image has not been enhanced.}. Notice the contrast has been improved significantly. Figure \ref{fig:rankillus} illustrates the more subtle effects of our ranking scheme. $r(x)$ is the rank ($1 \times 11$ mask) of $f(x)$ which is a sine wave with decaying mean, amplitude and frequency (equation \ref{eq:wave}).

\begin{equation}
\begin{array}{rcl}
f(x) & = & g(x \times \frac{2 \pi}{200}) \\
&&\\
g(z) & = & \frac{1}{2}(e^{- z / 2} + \frac{1}{2} e^{- z / 2} (1 + \sin(10z^{0.7}))) \\
\end{array}
\label{eq:wave}
\end{equation}

\subsection{Clustering}
The final process determines the location of the topic boundaries. The method is based on Reynar's maximisation algorithm \cite{reynar_1998,helfman_1996,church_1993,church_helfman_93}. A text segment is defined by two sentences $i,j$ (inclusive). This is represented as a square region along the diagonal of the rank matrix. Let $s_{i,j}$ denote the sum of the rank values in a segment and $a_{i,j} = (j-i+1)^2$ be the inside area. $B = \{b_1,...,b_m\}$ is a list of $m$ coherent text segments. $s_k$ and $a_k$ refers to the sum of rank and area of segment $k$ in $B$. $D$ is the inside density of $B$ (see equation \ref{eq:insidedensity}).

\begin{equation}
D = \frac{\sum_{k=1}^{m} s_k}{\sum_{k=1}^{m} a_k}
\label{eq:insidedensity}
\end{equation}

\begin{figure}[h]
\begin{center}
\includegraphics[width=0.48\textwidth]{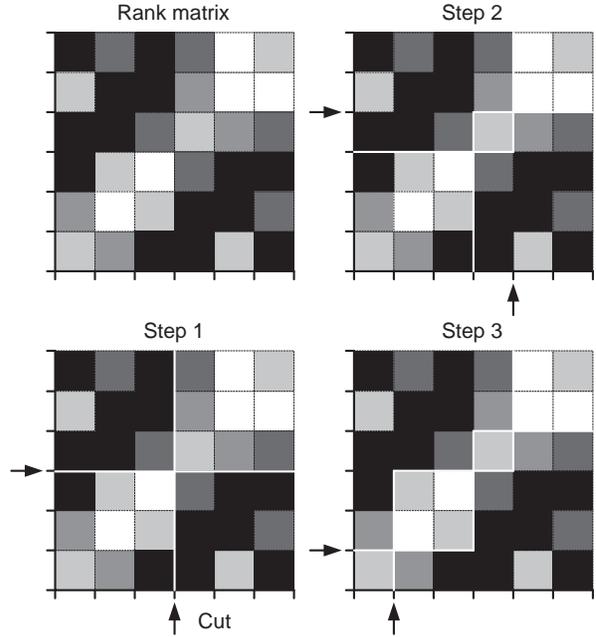}
\end{center}
\caption{A working example of the divisive clustering algorithm.}
\label{fig:split}
\end{figure}

To initialise the process, the entire document is placed in $B$ as one coherent text segment. Each step of the process splits one of the segments in $B$. The split point is a potential boundary which maximises $D$. Figure \ref{fig:split} shows a working example.

\begin{figure}[h]
\begin{center}
\includegraphics[width=0.48\textwidth]{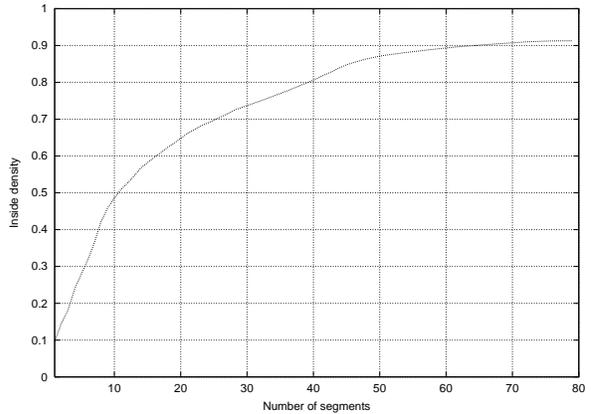}
\end{center}
\caption{The inside density of all levels of segmentation.}
\label{fig:densityPlot}
\end{figure}

\begin{figure}[h]
\begin{center}
\includegraphics[width=0.48\textwidth]{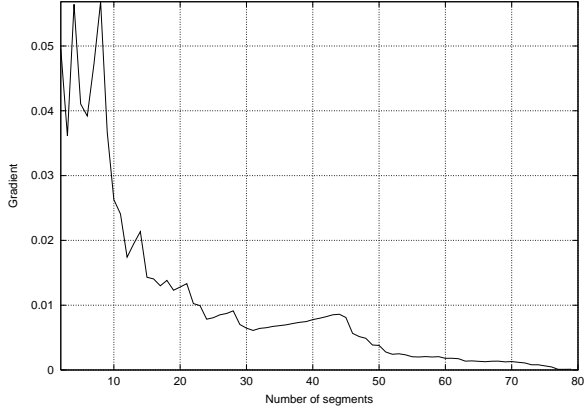}
\end{center}
\caption{Finding the optimal segmentation using the gradient.}
\label{fig:gradientPlot}
\end{figure}

The number of segments to generate, $m$, is determined automatically. $D^{(n)}$ is the inside density of $n$ segments and $\delta D^{(n)} = D^{(n)} - D^{(n-1)}$ is the gradient. For a document with $b$ potential boundaries, $b$ steps of divisive clustering generates $\{D^{(1)},...,D^{(b+1)}\}$ and $\{\delta D^{(2)},...,\delta D^{(b+1)}\}$ (see figure \ref{fig:densityPlot} and \ref{fig:gradientPlot}). An unusually large reduction in $\delta D$ suggests the optimal clustering has been obtained\footnote{In practice, convolution (mask $\{1,2,4,8,4,2,1\}$) is first applied to $\delta D$ to smooth out sharp local changes} (see $n=10$ in figure \ref{fig:gradientPlot}). Let $\mu$ and $\nu$ be the mean and variance of $\delta D^{(n)}, n \in \{2,...,b+1\}$. $m$ is obtained by applying the threshold, $\mu + c \times \sqrt{\nu}$, to $\delta D$ ($c=1.2$ works well in practice).

\subsection{Speed optimisation}
\label{sec:speed}
\begin{figure}[h]
\begin{center}
\includegraphics[width=0.48\textwidth]{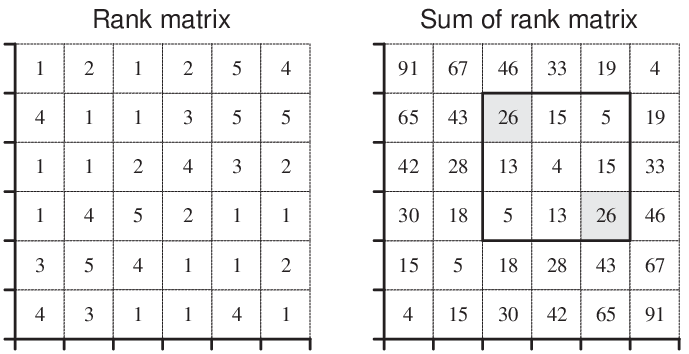}
\end{center}
\caption{Improving speed performance by pre-computing $s_{i,j}$.}
\label{fig:sumrank}
\end{figure}

The running time of each step is dominated by the computation of $s_k$. Given $s_{i,j}$ is constant, our algorithm pre-computes all the values to improve speed performance. The procedure computes the values along diagonals, starting from the main diagonal and works towards the corner. The method has a complexity of order $1\frac{1}{2} n^2$. Let $r_{i,j}$ refer to the rank value in the rank matrix $R$ and $S$ to the sum of rank matrix. Given $R$ of size $n \times n$, $S$ is computed in three steps (see equation \ref{eq:sumrank}). Figure \ref{fig:sumrank} shows the result of applying this procedure to the rank matrix in figure \ref{fig:split}.

\begin{equation}
\begin{array}{llcl}
1. & s_{i,i}   & = & r_{i,i} \\
   &           & \textrm{for}  & i \in \{1,...,n\} \\
2. & s_{i+1,i} & = & 2 r_{i+1,i} + s_{i,i} + s_{i+1,i+1} \\
   & s_{i,i+1} & = & s_{i+1,i} \\
   &           & \textrm{for} & i \in \{1,...,n-1\} \\
3. & s_{i+j,i} & = & 2 r_{i+j,i} + s_{i+j-1,i} +\\
   &           &   & s_{i+j,i+1} - s_{i+j-1,i+1} \\
   & s_{i,i+j} & = & s_{i+j,i} \\
   &           & \textrm{for} & j \in \{2,...,n-1\} \\
   &           &   & i \in \{1,...,n-j\} \\
\end{array}
\label{eq:sumrank}
\end{equation}

\section{Evaluation}
The definition of a topic segment ranges from complete stories \cite{allan_et_al_1998} to summaries \cite{ponte_croft_1997}. Given the quality of an algorithm is task dependent, the following experiments focus on the relative performance. Our evaluation strategy is a variant of that described in \cite[71-73]{reynar_1998} and the TDT segmentation task \cite{allan_et_al_1998}. We assume a good algorithm is one that finds the most prominent topic boundaries.
 
\subsection{Experiment procedure}
An artificial test corpus of 700 samples is used to assess the accuracy and speed performance of segmentation algorithms. A sample is a concatenation of ten text segments. A segment is the first $n$ sentences of a randomly selected document from the Brown corpus\footnote{Only the news articles {\tt ca**.pos} and informative text {\tt cj**.pos} were used in the experiment.}. A sample is characterised by the range of $n$. The corpus was generated by an automatic procedure\footnote{All experiment data, algorithms, scripts and detailed results are available from the author.}. Table \ref{tbl:testCorpus} presents the corpus statistics.

\begin{table}[ht]
\begin{center}
\begin{tabular}{|l|c|c|c|c|}
\hline
Range of $n$ & $3-11$ & $3-5$ & $6-8$ & $9-11$ \\
\hline
\# samples & 400 & 100 & 100 & 100 \\
\hline
\end{tabular}
\end{center}
\caption{Test corpus statistics.}
\label{tbl:testCorpus}
\end{table}

\begin{equation}
\begin{array}{l}
p(\textrm{error}|\textrm{ref},\textrm{hyp},k) = \\
\quad p(\textrm{miss}|\textrm{ref},\textrm{hyp},\textrm{diff},k)
p(\textrm{diff}|\textrm{ref},k) + \\
\quad p(\textrm{fa}|\textrm{ref},\textrm{hyp},\textrm{same},k)
p(\textrm{same}|\textrm{ref},k) \\
\end{array}
\label{eq:beefermanMetric}
\end{equation}

Speed performance is measured by the average number of CPU seconds required to process a test sample\footnote{All experiments were conducted on a Pentium II 266MHz PC with 128Mb RAM running RedHat Linux 6.0 and the Blackdown Linux port of JDK1.1.7 v3.}. Segmentation accuracy is measured by the error metric (equation \ref{eq:beefermanMetric}, fa $\to$ false alarms) proposed in \cite{beeferman_et_al_1999}. Low error probability indicates high accuracy. Other performance measures include the popular precision and recall metric (PR) \cite{hearst_1994}, fuzzy PR \cite{reynar_1998} and edit distance \cite{ponte_croft_1997}. The problems associated with these metrics are discussed in \cite{beeferman_et_al_1999}.

\subsection{Experiment 1 - Baseline}
Five degenerate algorithms define the baseline for the experiments. $B_n$ does not propose any boundaries. $B_a$ reports all potential boundaries as real boundaries. $B_e$ partitions the sample into regular segments. $B_{(r,?)}$ randomly selects any number of boundaries as real boundaries. $B_{(r,b)}$ randomly selects $b$ boundaries as real boundaries.

The accuracy of the last two algorithms are computed analytically. We consider the status of $m$ potential boundaries as a bit string ($1 \to$ topic boundary). The terms $p(\textrm{miss})$ and $p(\textrm{fa})$ in equation \ref{eq:beefermanMetric} corresponds to $p(\textrm{same}|k)$ and $p(\textrm{diff}|k) = 1 - p(\textrm{same}|k)$. Equation \ref{eq:bGen}, \ref{eq:bRanUnknown} and \ref{eq:bRanKnown} gives the general form of $p(\textrm{same}|k)$, $B_{(r,?)}$ and $B_{(r,b)}$, respectively\footnote{The full derivation of our method is available from the author.}. 

Table \ref{tbl:baseline} presents the experimental results. The values in row two and three, four and five are not actually the same. However, their differences are insignificant according to the Kolmogorov-Smirnov, or KS-test \cite{ks_test}.

\begin{equation}
p(\textrm{same}|k) = \frac{\textrm{\# valid segmentations}}{\textrm{\# possible segmentations}}
\label{eq:bGen}
\end{equation}

\begin{equation}
p(\textrm{same}|k,B_{(r,?)}) = \frac{2^{m-k}}{2^m} = 2^{-k}
\label{eq:bRanUnknown}
\end{equation}

\begin{equation}
\begin{array}{rcl}
p(\textrm{same}|k,m,B_{(r,b)}) & = & \frac{_{(m-k)}C_b}{_mC_b} \\
&&\\
_xC_y & = & \frac{x!}{y!(x-y)!} \\
\end{array}
\label{eq:bRanKnown}
\end{equation}
 
\begin{table}[ht]
\begin{center}
\begin{tabular}{|l|c|c|c|c|}
\hline
      & 3-11 & 3-5 & 6-8 & 9-11 \\
\hline
$B_e$ & 45\% & 38\% & 39\% & 36\% \\
$B_n$ & 47\% & 47\% & 47\% & 46\% \\
$B_{(r,b)}$ & 47\% & 47\% & 47\% & 46\% \\
$B_a$ & 53\% & 53\% & 53\% & 54\% \\
$B_{(r,?)}$ & 53\% & 53\% & 53\% & 54\% \\
\hline
\end{tabular}
\end{center}
\caption{The error rate of the baseline algorithms.}
\label{tbl:baseline}
\end{table}

\subsection{Experiment 2 - TextTiling}
We compare three versions of the TextTiling algorithm \cite{hearst_1994}. $H94_{(c,d)}$ is Hearst's C implementation with default parameters. $H94_{(c,r)}$ uses the recommended parameters $k=6$, $w=20$. $H94_{(j,r)}$ is my implementation of the algorithm. Experimental result (table \ref{tbl:textTile}) shows $H94_{(c,d)}$ and $H94_{(c,r)}$ are more accurate than $H94_{(j,r)}$. We suspect this is due to the use of a different stopword list and stemming algorithm.

\begin{table}[ht]
\begin{center}
\begin{tabular}{|l|c|c|c|c|}
\hline
      & 3-11 & 3-5 & 6-8 & 9-11 \\
\hline
$H94_{(c,d)}$ & 46\% & 44\% & 43\% & 48\% \\
$H94_{(c,r)}$ & 46\% & 44\% & 44\% & 49\% \\
$H94_{(j,r)}$ & 54\% & 45\% & 52\% & 53\% \\
\hline
$H94_{(c,d)}$ & 0.67s & 0.52s & 0.66s & 0.88s\\
$H94_{(c,r)}$ & 0.68s & 0.52s & 0.67s & 0.92s\\
$H94_{(j,r)}$ & 3.77s & 2.21s & 3.69s & 5.07s\\
\hline
\end{tabular}
\end{center}
\caption{The error rate and speed performance of TextTiling.}
\label{tbl:textTile}
\end{table}

\subsection{Experiment 3 - DotPlot}
Five versions of Reynar's optimisation algorithm \cite{reynar_1998} were evaluated. $R98$ and $R98_{(min)}$ are exact implementations of his maximisation and minimisation algorithm. $R98_{(s,cos)}$ is my version of the maximisation algorithm which uses the cosine coefficient instead of dot density for measuring similarity. It incorporates the optimisations described in section \ref{sec:speed}. $R98_{(m,dot)}$ is the modularised version of $R98$ for experimenting with different similarity measures.

$R98_{(m,sa)}$ uses a variant of Kozima's semantic similarity measure \cite{kozima_1993} to compute block similarity. Word similarity is a function of word co-occurrence statistics in the given document. Words that belong to the same sentence are considered to be related. Given the co-occurrence frequencies $f(w_i,w_j)$, the transition probability matrix $t$ is computed by equation \ref{eq:trans}. Equation \ref{eq:spread} defines our spread activation scheme. $s$ denotes the word similarity matrix, $x$ is the number of activation steps and $\textrm{norm}(y)$ converts a matrix $y$ into a transition matrix. $x=5$ was used in the experiment.

\begin{equation}
t_{i,j} = p(w_j|w_i) = \frac{f(w_i,w_j)}{\sum_j f(w_i,w_j)}
\label{eq:trans}
\end{equation}

\begin{equation}
s = \textrm{norm} \left(\sum_{i=1}^{x} t^i \right)
\label{eq:spread}
\end{equation}

Experimental result (table \ref{tbl:dotplot}) shows the cosine coefficient and our spread activation method improved segmentation accuracy. The speed optimisations significantly reduced the execution time.

\begin{table}[ht]
\begin{center}
\begin{tabular}{|l|c|c|c|c|}
\hline
 & 3-11 & 3-5 & 6-8 & 9-11\\
\hline
$R98_{(m,sa)}$ & 18\% & 20\% & 15\% & 12\% \\
$R98_{(s,cos)}$ & 21\% & 18\% & 19\% & 18\% \\
$R98_{(m,dot)}$ & 22\% & 21\% & 18\% & 16\% \\
$R98$ & 22\% & 21\% & 18\% & 16\% \\
$R98_{(min)}$ & n/a & 34\% & 37\% & 37\% \\
\hline
$R98_{(s,cos)}$ & 4.54s & 2.24s & 4.36s & 6.99s\\
$R98$ & 29.58s & 9.29s & 28.09s & 55.03s\\
$R98_{(m,sa)}$ & 41.02s & 7.34s & 40.05s & 113.5s\\
$R98_{(m,dot)}$ & 46.58s & 9.24s & 42.72s & 115.4s\\
$R98_{(min)}$ & n/a & 19.62s & 58.77s & 122.6s\\
\hline
\end{tabular}
\end{center}
\caption{The error rate and speed performance of Reynar's optimisation algorithm.}
\label{tbl:dotplot}
\end{table}

\subsection{Experiment 4 - Segmenter}
We compare three versions of Segmenter \cite{kan_et_al_1998}. $K98_{(p)}$ is the original Perl implementation of the algorithm (version 1.6). $K98_{(j)}$ is my implementation of the algorithm. $K98_{(j,a)}$ is a version of $K98_{(j)}$ which uses a document specific chain breaking strategy. The distribution of link distances are used to identify unusually long links. The threshold is a function $\mu + c \times \sqrt{\nu}$ of the mean $\mu$ and variance $\nu$. We found $c = 1$ works well in practice.

Table \ref{tbl:segmenter} summarises the experimental results. $K98_{(p)}$ performed significantly better than $K98_{(j,*)}$. This is due to the use of a different part-of-speech tagger and shallow parser. The difference in speed is largely due to the programming languages and term clustering strategies. Our chain breaking strategy improved accuracy (compare $K98_{(j)}$ with $K98_{(j,a)}$).

\begin{table}[ht]
\begin{center}
\begin{tabular}{|l|c|c|c|c|}
\hline
 & 3-11 & 3-5 & 6-8 & 9-11\\
\hline
$K98_{(p)}$ & 36\% & 23\% & 33\% & 43\% \\
$K98_{(j,a)}$ & n/a & 41\% & 46\% & 50\% \\
$K98_{(j)}$ & n/a & 44\% & 48\% & 51\% \\
\hline
$K98_{(p)}$ & 4.24s & 2.57s & 4.21s & 6.00s \\
$K98_{(j)}$ & n/a & 21.43s & 65.54s & 129.3s \\
$K98_{(j,a)}$ & n/a & 21.44s & 65.49s & 129.7s \\
\hline
\end{tabular}
\end{center}
\caption{The error rate and speed performance of Segmenter.}
\label{tbl:segmenter}
\end{table}

\subsection{Experiment 5 - Our algorithm, $C99$}
Two versions of our algorithm were developed, $C99$ and $C99_{(b)}$. The former is an exact implementation of the algorithm described in this paper. The latter is given the expected number of topic segments for fair comparison with $R98$. Both algorithms used a $11 \times 11$ ranking mask.

The first experiment focuses on the impact of our automatic termination strategy on $C99_{(b)}$ (table \ref{tbl:c99}). $C99_{(b)}$ is marginally more accurate than $C99$. This indicates our automatic termination strategy is effective but not optimal. The minor reduction in speed performance is acceptable.

\begin{table}[ht]
\begin{center}
\begin{tabular}{|l|c|c|c|c|}
\hline
 & 3-11 & 3-5 & 6-8 & 9-11\\
\hline
$C99_{(b)}$ & 12\% & 12\% & 9\% & 9\% \\
$C99$ & 13\% & 18\% & 10\% & 10\% \\
\hline
$C99_{(b)}$ & 4.00s & 1.91s & 3.73s & 5.99s \\
$C99$ & 4.04s & 2.12s & 4.04s & 6.31s \\
\hline
\end{tabular}
\end{center}
\caption{The error rate and speed performance of our algorithm, $C99$.}
\label{tbl:c99}
\end{table}

The second experiment investigates the effect of different ranking mask size on the performance of $C99$ (table \ref{tbl:c99mask}). Execution time increases with mask size. A $1 \times 1$ ranking mask reduces all the elements in the rank matrix to zero. Interestingly, the increase in ranking mask size beyond $3 \times 3$ has insignificant effect on segmentation accuracy. This suggests the use of extrema for clustering has a greater impact on accuracy than linearising the similarity scores (figure \ref{fig:rankillus}).

\begin{table}[ht]
\begin{center}
\begin{tabular}{|l|c|c|c|c|}
\hline
 & 3-11 & 3-5 & 6-8 & 9-11\\
\hline
$1 \times 1$ & 48\% & 48\% & 50\% & 49\% \\
$3 \times 3$ & 12\% & 11\% & 10\% & 8\% \\
$5 \times 5$ & 12\% & 11\% & 10\% & 8\% \\
$7 \times 7$ & 12\% & 11\% & 10\% & 8\% \\
$9 \times 9$ & 12\% & 11\% & 10\% & 9\% \\
$11 \times 11$ & 12\% & 11\% & 10\% & 9\% \\
$13 \times 13$ & 12\% & 11\% & 10\% & 9\% \\
$15 \times 15$ & 12\% & 11\% & 10\% & 9\% \\
$17 \times 17$ & 12\% & 10\% & 10\% & 8\% \\
\hline
$1 \times 1$ & 3.92s & 2.06s & 3.84s & 5.91s\\
$3 \times 3$ & 3.83s & 2.03s & 3.79s & 5.85s\\
$5 \times 5$ & 3.86s & 2.04s & 3.84s & 5.92s\\
$7 \times 7$ & 3.90s & 2.06s & 3.88s & 6.00s\\
$9 \times 9$ & 3.96s & 2.07s & 3.92s & 6.12s\\
$11 \times 11$ & 4.02s & 2.09s & 3.98s & 6.26s\\
$13 \times 13$ & 4.11s & 2.11s & 4.07s & 6.41s\\
$15 \times 15$ & 4.20s & 2.14s & 4.14s & 6.60s\\
$17 \times 17$ & 4.29s & 2.17s & 4.25s & 6.79s\\
\hline
\end{tabular}
\end{center}
\caption{The impact of mask size on the performance of $C99$.}
\label{tbl:c99mask}
\end{table}

\subsection{Summary}
Experimental result (table \ref{tbl:sum}) shows our algorithm $C99$ is more accurate than existing algorithms. A two-fold increase in accuracy and seven-fold increase in speed was achieved (compare $C99_{(b)}$ with $R98$). If one disregards segmentation accuracy, $H94$ has the best algorithmic performance (linear). $C99$, $K98$ and $R98$ are all polynomial time algorithms. The significance of our results has been confirmed by both t-test and KS-test.

\begin{table}[ht]
\begin{center}
\begin{tabular}{|l|c|c|c|c|}
\hline
      & 3-11 & 3-5 & 6-8 & 9-11 \\
\hline
$C99_{(b)}$ & 12\% & 12\% & 9\% & 9\% \\
$C99$ & 13\% & 18\% & 10\% & 10\% \\
$R98$ & 22\% & 21\% & 18\% & 16\% \\
$K98_{(p)}$ & 36\% & 23\% & 33\% & 43\% \\
$H94_{(c,d)}$ & 46\% & 44\% & 43\% & 48\% \\
\hline
$H94_{(j,r)}$ & 3.77s & 2.21s & 3.69s & 5.07s\\
$C99_{(b)}$ & 4.00s & 1.91s & 3.73s & 5.99s \\
$C99$ & 4.04s & 2.12s & 4.04s & 6.31s \\
$R98$ & 29.58s & 9.29s & 28.09s & 55.03s\\
$K98_{(j)}$ & n/a & 21.43s & 65.54s & 129.3s \\
\hline
\end{tabular}
\end{center}
\caption{A summary of our experimental results.}
\label{tbl:sum}
\end{table}

\section{Conclusions and future work}
A segmentation algorithm has two key elements, a clustering strategy and a similarity measure. Our results show divisive clustering ($R98$) is more precise than sliding window ($H94$) and lexical chains ($K98$) for locating topic boundaries.

Four similarity measures were examined. The cosine coefficient ($R98_{(s,cos)}$) and dot density measure ($R98_{(m,dot)}$) yield similar results. Our spread activation based semantic measure ($R98_{(m,sa)}$) improved accuracy. This confirms that although Kozima's approach \cite{kozima_1993} is computationally expensive, it does produce more precise segmentation.

The most significant improvement was due to our ranking scheme which linearises the cosine coefficient. Our experiments demonstrate that given insufficient data, the qualitative behaviour of the cosine measure is indeed more reliable than the actual values.

Although our evaluation scheme is sufficient for this comparative study, further research requires a large scale, task independent benchmark. It would be interesting to compare $C99$ with the multi-source method described in \cite{beeferman_et_al_1999} using the TDT corpus. We would also like to develop a linear time and multi-source version of the algorithm.

\section*{Acknowledgements}
This paper has benefitted from the comments of Mary McGee Wood and the anonymous reviewers. Thanks are due to my parents and department for making this work possible; Jeffrey Reynar for discussions and guidance on the segmentation problem; Hideki Kozima for help on the spread activation measure; Min-Yen Kan and Marti Hearst for their segmentation algorithms; Daniel Oram for references to image processing techniques; Magnus Rattray and Stephen Marsland for help on statistics and mathematics.

\bibliographystyle{acl}
\bibliography{references}
\end{document}